\documentclass[journal]{IEEEtran}
\usepackage{cite}

\ifCLASSINFOpdf
\else
\fi
\usepackage{amsmath}
\DeclareMathOperator*{\argminA}{arg\,min}
\DeclareMathOperator*{\argmaxA}{arg\,max}
\usepackage{array}
\usepackage{url}

\usepackage{graphics}
\usepackage{graphicx}
\usepackage{epstopdf}
\usepackage{amssymb}
\usepackage{algorithm,algorithmic}
\usepackage{hyperref}
\usepackage{url}
\usepackage{multirow}
\usepackage{array}
\newcolumntype{M}[1]{>{\centering\arraybackslash}m{#1}}
\usepackage{amsmath}
\usepackage{float}
\usepackage[caption=false]{subfig}
\usepackage{color}
\usepackage{caption}
\usepackage{dsfont}

\begin{document}
%
\title{Unsupervised Domain Adaptation for Cross-sensor Pore Detection in  High-resolution Fingerprint Images}
%
%
%

\author{Vijay~Anand,~\IEEEmembership{Student Member,~IEEE,}
        and~Vivek~Kanhangad,~\IEEEmembership{Senior Member,~IEEE}
\thanks{V. Anand  and V. Kanhangad are with the Discipline of Electrical Engineering, Indian Institute of Technology Indore, Indore 453552, India
 e-mail: phd1401202011@iiti.ac.in (V. Anand),  kvivek@iiti.ac.in (V. Kanhangad).}
}

\IEEEspecialpapernotice{(``This work has been submitted to the IEEE for possible publication. Copyright may be transferred without notice, after which this version may no longer be accessible.")}

\maketitle

\begin{abstract}
With the emergence of high-resolution fingerprint sensors, there has been a lot of focus on level-3 fingerprint features, especially the pores, for the next generation automated fingerprint recognition systems (AFRS). Following the success of deep learning in various computer vision tasks, researchers have developed learning-based approaches for detection of pores in high-resolution fingerprint images.
Generally, learning-based approaches provide better performance than hand-crafted feature-based approaches.
However, domain adaptability of the existing learning-based pore detection methods has never been studied. In this paper, we study this aspect and propose an approach for pore detection in cross-sensor scenarios. For this purpose, we have generated an in-house 1000 dpi fingerprint dataset with ground truth pore coordinates (referred to as IITI-HRFP-GT), and evaluated the performance of the existing learning-based pore detection approaches. The core of the proposed approach for detection of pores in cross-sensor scenarios is DeepDomainPore, which is a   residual learning-based convolutional neural network (CNN) trained for pore detection.   The domain adaptability in DeepDomainPore is achieved by embedding a gradient reversal layer between the CNN and a domain classifier network. The proposed approach achieves state-of-the-art performance in a cross-sensor scenario involving public high-resolution fingerprint datasets with 88.12\% true detection rate and 83.82\% F-score.  
\end{abstract}


%
\IEEEpeerreviewmaketitle

\section{Introduction}
\label{intro}
%
%
%
%
\IEEEPARstart{L}{evel-3} fingerprint features are generally observable in high-resolution fingerprint images of resolution higher than 800 dpi \cite{HEF_resolution}. 
With advancements in fingerprint sensing technology, researchers have focused their attention on level-3 fingerprint features, especially the pores and proposed various matching techniques for next generation automated fingerprint recognition systems (AFRS) \cite{stosz_pore,roddy1997fingerprint,kryszczuk2004extraction,kryszczuk2004study,jain2007pores, zhao2009direct, Zhao20102833, ZHAO_partial,sparse_fing, LIU_PR,Lemes,segundo,vijay_pore}. Pore features have been found to carry sufficient discriminating power and shown to be effective for biometric recognition using even partial fingerprint images, which may not always contain sufficient level-2 features \cite{zhao2009direct}. 
However, it is imperative that the pore coordinates are detected accurately as the pore detector hugely impacts the performance of pore-based AFRS.  The existing pore detection approaches can be broadly classified into feature-based approaches \cite{stosz_pore,roddy1997fingerprint,kryszczuk2004extraction,kryszczuk2004study,jain2007pores, zhao2009direct, Zhao20102833, ZHAO_partial,sparse_fing, LIU_PR,Lemes,segundo} and the learning-based approaches \cite{LABATI2017,Deep_pore,vij_cnn,Dahia_pore_detection}.

The early feature-based pore detection approaches \cite{stosz_pore,roddy1997fingerprint,kryszczuk2004extraction,kryszczuk2004study} performed skeletonization on very high-resolution ($\sim$ 2000 dpi) fingerprint images. Their performance is likely to be adversely affected by fingerprint degradation caused by skin conditions.
 To overcome this drawback, Jain \textit{et al.} \cite{jain2007pores} presented a pore detection approach that  applies Mexican-hat wavelet transform to the linear combination of original and enhanced fingerprint image. 
 However, their approach does not possess adaptive capacity to handle changes in size of the pores. As a result, it suffers from high false pore detection.
Later on, Zhao \textit{et al.} \cite{zhao2009direct} proposed an adaptive filtering method \cite{zhao_ICPR} for pore detection.
The authors have also demonstrated the usefulness of pores for biometric recognition using partial fingerprint images, which may not contain sufficient level-2 features \cite{Zhao20102833,ZHAO_partial}. 
Teixeria and Leite \cite{Teixeira} presented an improved pore detection method that relies on the analysis of spatial distribution of the detected pores. Pores are detected by applying morphological scale-space toggle operator in combination with the directional field information of the fingerprint image. The detected pores are then classified as true positive or false positive pores based on the distance between adjacent pores and the average width of the corresponding ridge.
The approaches presented in \cite{zhao2009direct,Teixeira} improved the pore detection accuracy, but at the cost of increased computational complexity.  
Lemes \textit{et al.} \cite{Lemes} proposed a dynamic pore filtering approach with a low computational cost.  
Their approach is adaptive and handles variations in the pore size. 
Later on, Segundo and Lemes \cite{segundo} improved the pore detection approach presented in \cite{Lemes} by employing the average ridge width in place of the average valley width to obtain the global and local radii, which  are  used  in  the  same  manner  as  in  \cite{Lemes}  to  estimate the  pore  coordinates.

Following the success of deep learning in various computer vision tasks \cite{facenet,Deepface,deep_conv}, researchers have focused on developing learning-based pore detection approaches.
In their pioneering work, Labati \emph{et al.} \cite{LABATI2017} proposed an approach for pore detection using a shallow CNN that generates a pore intensity map, which is thresholded to obtain the pore coordinates. However, the approach \cite{LABATI2017} does not provide any improvement in performance over the existing feature-based techniques. Later on, Jang \emph{et al.}\cite{Deep_pore} presented a pore detection approach that employs a deep CNN consisting of 10 learnable layers. In their approach,  pore label image is generated by considering the distance of every pixel from the ground truth pore coordinates. The trained deep CNN generates a pore intensity map, which is further processed to obtain the pore coordinates. DeepPore \cite{Deep_pore} provides a considerable improvement in pore detection accuracy over the existing approaches. However, DeepPore \cite{Deep_pore} has a plain CNN architecture with only ten learnable layers and has been tested on a very small test set containing only six fingerprint images.
Authors in \cite{vij_cnn} proposed a residual learning-based CNN, referred to as DeepResPore, to detect pores in high-resolution fingerprint images. DeepResPore model contains 18 layers with eight residual blocks. It has been trained on a large labeled dataset containing 210,330 patches and evaluated on two test sets, each containing 30 fingerprint images to ascertain its performance. DeepResPore-based approach \cite{vij_cnn} provides state-of-the-art performance on both the test sets.
{Dahia and Segundo \cite{Dahia_pore_detection} employed a fully convolutional neural network (FCN) to detect pores in high-resolution fingerprint images. They have also presented a protocol to evaluate a pore detector.}

A review of the literature indicates that learning-based pore detection approaches generally perform better than the feature-based pore detection approaches.  However, all of the existing learning-based pore detection approaches have been evaluated only on fingerprint images from the benchmark PolyU HRF datasets \cite{POLYU}. Specifically, these approaches have used the ground truth of pore coordinates provided in the PolyU HRF dataset, which contains high-resolution fingerprint images of resolution 1200 dpi captured using a single sensor.
Thus, their domain adaptability in a cross-sensor scenario has never been studied. The  reason  could be that ground truth pore  data for high-resolution  fingerprint  images acquired using a  different sensor is unavailable. The ground truth creation is not only a challenging but also a time-consuming process as coordinates of the pores present in a fingerprint image need to marked manually.  
The aforementioned challenges motivated us to evaluate existing learning-based pore detection approaches on cross-sensor high-resolution fingerprint images and develop a learning-based pore detection approach suited for \textit{cross-sensor scenario, without requiring any labeled pore coordinates for high-resolution fingerprints from the new dataset}. To the best of our knowledge, this is the first study that investigates domain adaptability of learning-based pore detection approaches on cross-sensor fingerprint images.

In this work, we have generated an in-house 1000 dpi fingerprint dataset with ground truth pore coordinates, collectively referred to as IITI-HRFP-GT. 
Specifically, IITI-HRFP-GT contains 6,406 annotated pores with each of their coordinates marked manually. The commercially available Biometrika HiScan-Pro fingerprint scanner has been employed for fingerprint image acquisition.

The key contributions of this paper are as follows:  we present results of the first study of domain adaptability of the existing learning-based pore detection approaches on cross-sensor fingerprint images. The in-house dataset used in this study will be made available to further research in this area. 
Most importantly, this paper presents an approach to detect pores in cross-sensor scenarios using unsupervised domain adaptation technique. In the proposed approach, 
 domain adaptability is achieved by incorporating a gradient reversal layer during the training phase.  

The rest of this paper is organized as follows: Section \ref{PM} presents the proposed methodology. It also describes in detail how the proposed CNN model is trained in a cross-sensor scenario. 
Experimental results and discussion are presented in Section \ref{results}, which also presents a detailed description of the dataset preparation for cross-sensor evaluation of the proposed model. Finally, Section \ref{conclude} presents our concluding remarks. 

\section{Proposed Method}
\label{PM}
\begin{figure*}[!ht]
\centering
{\includegraphics[width=0.9\textwidth]{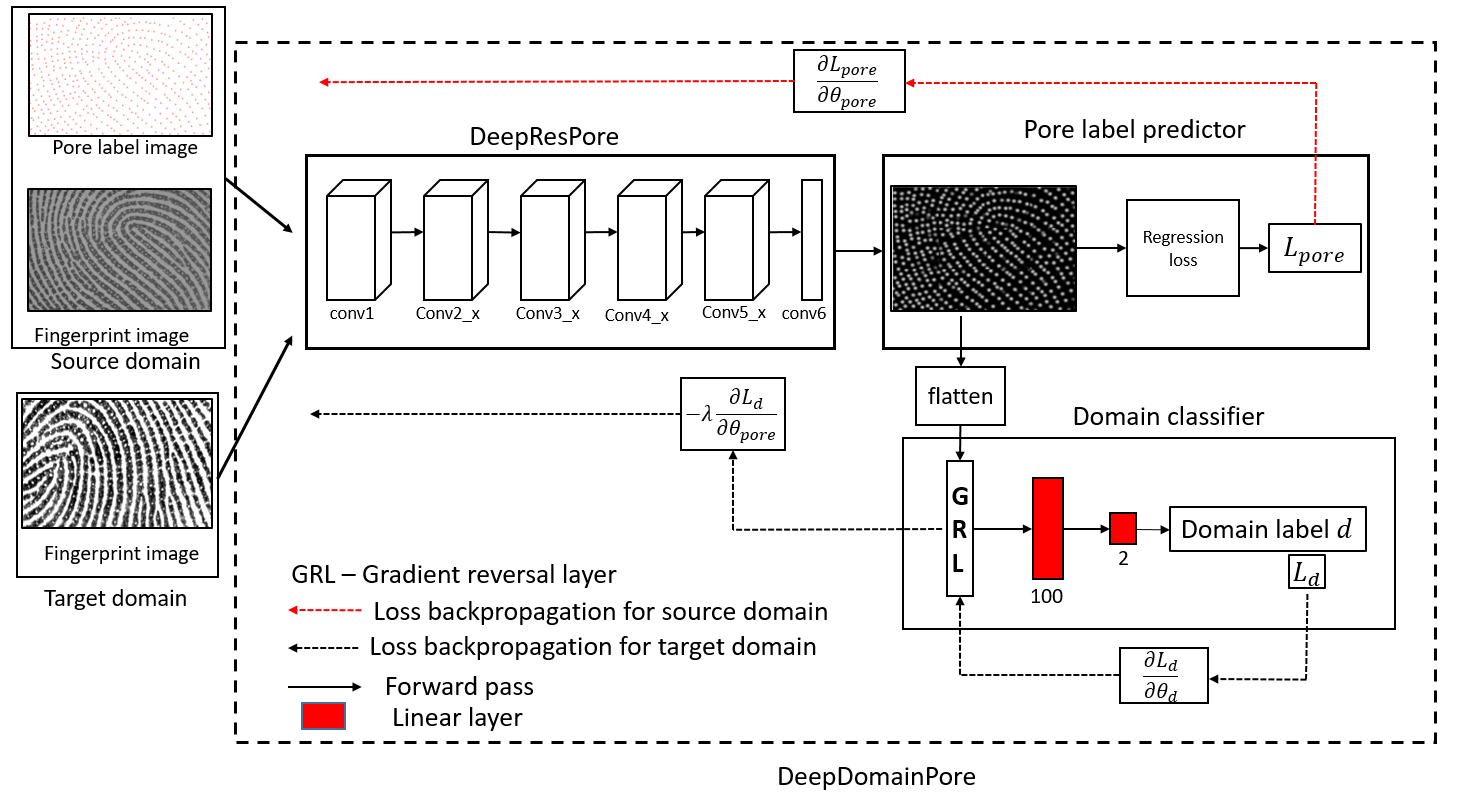}}
\caption{Schematic diagram of the DeepDomainPore model}
\label{deepdomainpore}
\end{figure*}

In this paper, we present an automated approach for pore detection in cross-sensor scenarios. Specifically, the proposed pore detection approach employs a customized CNN model, referred to as DeepDomainPore. 
DeepDomainPore is a combination of a residual learning-based CNN model DeepResPore \cite{vij_cnn} and an unsupervised domain adaptation approach, incorporated by embedding a  gradient reversal layer while training DeepResPore model. The proposed DeepDomainPore is trained using labeled data from the source domain and unlabeled data from the target domain. A schematic diagram of the proposed approach is presented in Fig. \ref{deepdomainpore}. 

\begin{figure*}[!ht]
\centering
{\includegraphics[width=0.9\textwidth]{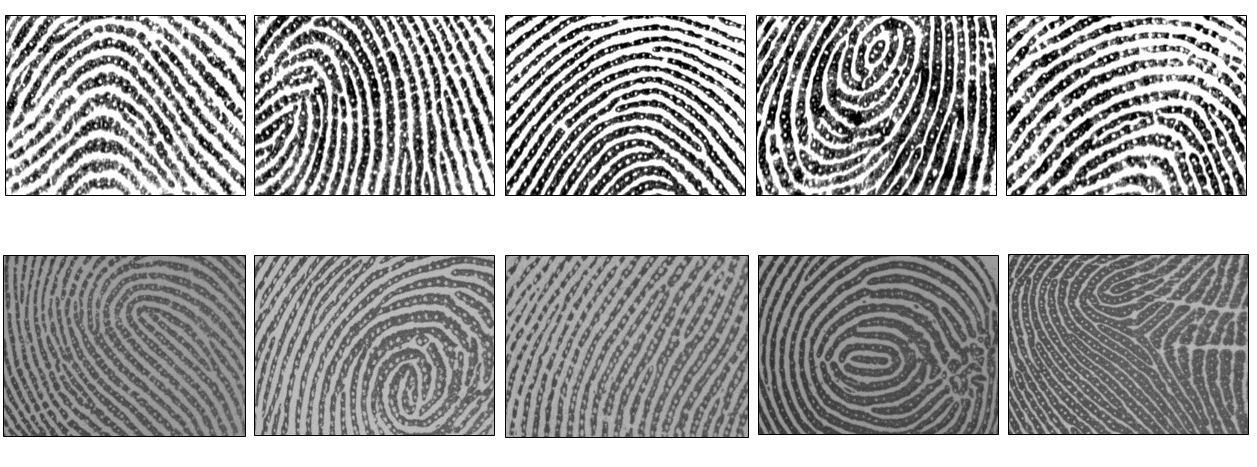}}
\caption{Sample fingerprint images from IITI-HRFP (top row) and PolyU HRF (bottom row)}
\label{sample_img}
\end{figure*}

The domain adaptation is defined as the process of learning a discriminative classifier or a predictor in the presence of a shift between the training and the test distribution \cite{Deep_DA, Ganin_DANN}. The proposed learning-based pore detection approach incorporates  domain adaptation during the training process. The need for domain adaption is illustrated in Fig. \ref{sample_img}, which presents sample fingerprint images from two high-resolution fingerprint datasets namely, PolyU HRF \cite{POLYU} and IITI-HRFP \cite{Anand2019}. As can be observed, there exists a considerable difference in terms of the resolution and pore size between the fingerprint images belonging to PolyU HRF and IITI-HRFP datasets. The proposed pore detection approach is designed to handle such domain shifts typically encountered in cross-sensor scenarios.

To train DeepDomainPore model, we consider one of the datasets to be the source domain $S$. $S$ should consist of fingerprint patches $x^{s}_{i}$ and the corresponding pore label images $y_{i}$, which highlight coordinates of each of the pores and its neighbourhood.
On the other hand, the target domain $T$ contains only unlabeled fingerprint patches $x^{t}_{i}$.  

\begin{figure*}[!ht]
\centering
{\includegraphics[width=0.9\textwidth]{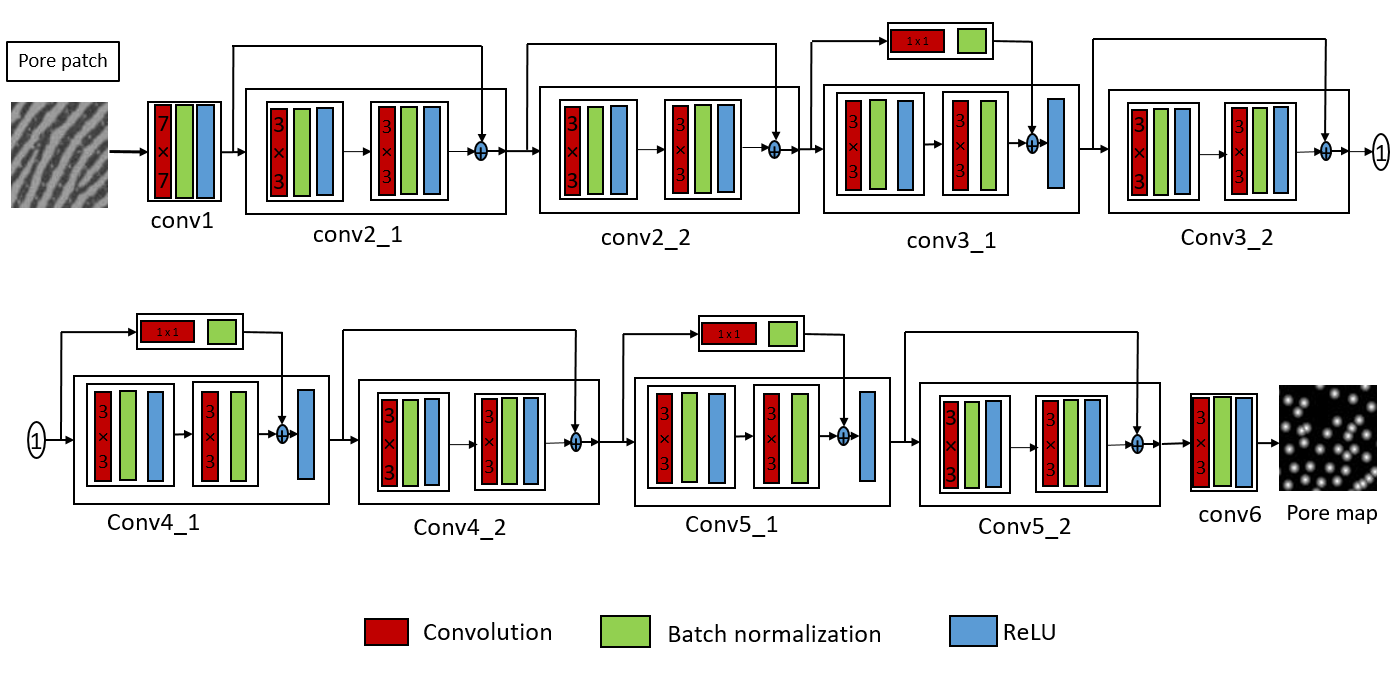}}
\caption{Schematic diagram of DeepResPore showing generation of pore map from a given fingerprint patch}
\label{deeprespore}
\end{figure*}

During training, fingerprint patches from both $S$ and $T$ are utilized. Each training sample $x_{i}$ is assigned a domain label $d_{i}$ in the following manner:
\begin{equation}
  d_{i}=\begin{cases}
    0, & \text{if $x_{i}\in S$}\\
    1, & \text{if $x_{i}\in T$}
  \end{cases}
\end{equation}
Thus, each training sample from $S$ has a pore label and a domain label. On the other hand, a training sample from $T$ only has its domain label.
As shown in Fig. \ref{deepdomainpore}, DeepDomainPore consists of DeepResPore \cite{vij_cnn} which is common between pore label predictor and the domain classifier. We have included an additional batch normalization (BN) and rectified linear unit (ReLU) layer succeeding the last convolutional layer of the DeepResPore model. The architecture of the DeepResPore is summarized in Table \ref{deep_res_pore_CNN}. As can be observed, the output size of each layer is the same as that of the input. We have performed padding to achieve this. Further, we have not employed any pooling operation in the network. 
A schematic diagram of the modified DeepResPore model is presented in Fig. \ref{deeprespore}. As depicted in Fig. \ref{deeprespore}, the modified DeepResPore model contains eight residual blocks, and additional BN and ReLU layers with the last convolutional layer (conv6).     

\begin{table}[h!]
\centering
\caption{Architecture of DeepResPore}
\label{deep_res_pore_CNN}
\begin{tabular}{|c|c|c|}
\hline
Layer name & output size & kernel                                                                        \\ \hline
conv1      & $80\times80$       & $7\times7$, $64$, stride 1, padding same                                                            \\ \hline
conv2\_x   &$80\times80$      & \Big[\begin{tabular}[c]{@{}c@{}}$3\times3$, $64$ \\ $3\times3$, $64$\end{tabular}\Big] $\times2$                  \\ \hline
conv3\_x   & $80\times80$      & \Big[\begin{tabular}[c]{@{}c@{}}$3\times3$, $128$ \\ $3\times3$, $128$\end{tabular}\Big] $\times2$                    \\ \hline
conv4\_x   &$80\times80$      & \Big[\begin{tabular}[c]{@{}c@{}}$3\times3$, $256$ \\ $3\times3$, $256$\end{tabular}\Big]  $\times2$                            \\ \hline
conv5\_x   &$80\times80$      & \Big[\begin{tabular}[c]{@{}c@{}}$3\times3$, $512$\\  $3\times3$, $512$\end{tabular}\Big]   $\times2$                           \\ \hline
conv6      & $80\times80$       & $3\times3$, $1$                                                                            \\ \hline
\end{tabular}
\end{table}

DeepResPore generates  pore intensity maps $\hat y$ of the size same as that of the input fingerprint patches $x_{i}$. Let this mapping and the parameters ($\theta_{pore}$) associated with it be represented as: 
\begin{equation}
    \hat y = f_{P}(X, \theta_{pore})
\end{equation}
where $X$ is the set of training fingerprint images.
The  pore intensity maps $\hat y$ are then compared with the corresponding pore label images $y$ to compute the loss $L_{pore}$ in the following way:
\begin{equation}\label{}
  L_{pore}=\frac{1}{n}\sum_{i=1}^{n}(y_{i}-\hat{y_{i}})^{2}
\end{equation}

For domain adaptation, the generated pore intensity map $\hat y$ is mapped to the domain label $\hat d \in \{0, 1 \}$  by a domain classifier function $f_{D}$. This mapping can be represented mathematically as follows:
\begin{equation}
\label{d}
    \hat d = f_{D}(X_{d}, \theta_{d})
\end{equation}
where $X_{d}$ is obtained  by flattening $\hat y$ and $\theta_{d}$ represents the parameters associated with the mapping in (\ref{d}).
It may be noted that the proposed DeepDomainPore model predicts both the pore label and the domain label for a fingerprint image from the source domain. On the other hand, it predicts only the domain label for a fingerprint image from the target domain.  
In the learning phase, the goal is to minimize the pore-label loss $L_{pore}$ on the annotated fingerprint images $x_{i}^{s}$ belonging to the source domain $S$. The set of parameters $\theta_{pore}$ of the  DeepResPore network is optimized to minimize $L_{pore}$ for fingerprint images from the source domain. Simultaneously, our approach aims to make the predictions $\hat y$ from DeepResPore model domain invariant. To this end, we have incorporated domain adaptation \cite{Ganin_DANN} in DeepResPore model. The resulting DeepDomainPore contains a domain classifier $f_{D}$, which takes a pore intensity map $\hat y$ 
generated by $f_{P}$ and employs a series of linear functions in combination with the softmax function to estimate the domain probability.
Thus, during training, our objective is to obtain a set of parameters $\theta_{pore}$ of the pore intensity map generation function $f_{P}$ that
minimizes the predicted pore loss $L_{pore}$ and maximizes the domain classifier loss  $L_{d}$, to ensure domain invariability in pore maps generated from the source and target domain fingerprint images. 
Further, we seek a set of domain classifier parameters $\theta_{d}$ that minimizes the domain classifier loss. 
To meet the aforementioned objectives, the following loss function is formulated:
\begin{multline}
\label{final_loss}
    E(\theta_{pore},\theta_{d}) =\frac{1}{n} \sum_{i=1}^{n} L_{pore}^{i,s}(\theta_{pore})-\lambda \bigg[\frac{1}{n}\sum_{i=1}^{n}L_{d}^{i,s}(\theta_{pore},\theta_{d})\\ +\frac{1}{n^{'}}\sum_{i=n+1}^{N}L_{d}^{i,t}(\theta_{pore},\theta_{d})\bigg]
\end{multline}
where $L_{pore}^{i,s}$ and $L_{d}^{i,s}$ represent predicted pore loss and domain classifier loss, respectively, for fingerprint images from the source domain. $L_{d}^{i,t}$ represents domain classifier loss for fingerprint images from the target domain. The parameter $\lambda$ controls the trade-off between the two objectives during the training.
In order to optimize (\ref{final_loss}), we need to find the saddle point $\hat{\theta}_{pore}$, $\hat{\theta}_{d}$ in such a way that:
\begin{equation}
\label{theta_pore}
 \hat{\theta}_{pore} = \argminA_{\theta_{pore}} E(\theta_{pore},\hat{\theta}_{d})    
\end{equation}

\begin{equation}
\label{theta_d}
 \hat{\theta}_{d} = \argmaxA_{\theta_{d}} E(\hat{\theta}_{pore},{\theta}_{d})     
\end{equation}

The saddle point defined in (\ref{theta_pore})-(\ref{theta_d}) can be obtained by computing the
stationary point of the gradient updates of $\theta_{pore}$ and $\theta_{d}$, as shown below:
\begin{equation}
\label{update1}
 \theta_{pore} \longleftarrow \theta_{pore} - \mu \bigg( \frac{\partial L_{pore}^{i}}{\partial \theta_{pore}}  - \lambda  \frac{\partial L_{d}^{i}}{\partial \theta_{pore}} \bigg)  
\end{equation}

\begin{equation}
\label{update2}
\theta_{d} \longleftarrow \theta_{d} - \mu\lambda \frac{\partial L_{d}^{i}}{\partial \theta_{d}}  
\end{equation}
where $\mu$ is the learning rate.
The updates in  (\ref{update1}) can be achieved by introducing a special layer termed \textit{gradient reversal layer} \cite{Ganin_DANN}. The gradient reversal layer has a hyper-parameter $\lambda$ (domain adaptation factor) associated with it. During the forward pass, the gradient reversal layer acts as an identity transform. On the other hand, during backpropagation, the gradient from the following layer is multiplied by $-\lambda$ and sent to the preceding layer. Mathematically, the forward and the backpropagation of the gradient reversal layer can be represented by the following pseudo-functions \cite{Ganin_DANN}:

\begin{equation}
    R_{\lambda}(x)=x
\end{equation}
\begin{equation}
    \frac{dR_{\lambda}}{dx}=-{\lambda}I
\end{equation}
where $I$ is an identity matrix.
In the proposed approach, we have included the gradient reversal layer between DeepResPore model $f_{P}$ and the domain classifier $f_{D}$, as shown in Fig. \ref{deepdomainpore}. During the backpropagation pass through the gradient reversal layer, partial derivative of the domain classifier loss $L_{d}$ with respect to the layer parameter  $\theta_{pore}$ \textit{i.e.} $\frac{\partial L_{d}}{\partial \theta_{pore}}$ is multiplied by $-\lambda$ and then passed to DeepResPore model.

In the test phase, a test fingerprint image is first divided into non-overlapping patches of size $80\times80$ pixels. These patches are processed individually by DeepDomainPore model to generate the corresponding pore intensity maps (of the same size as that of input patch), in which the pore locations are indicated by blobs. These pore intensity maps are then combined to form a complete pore map of the same size as that of the input fingerprint image.

\begin{figure*}[!ht]
\centering
{\includegraphics[width=0.9\textwidth]{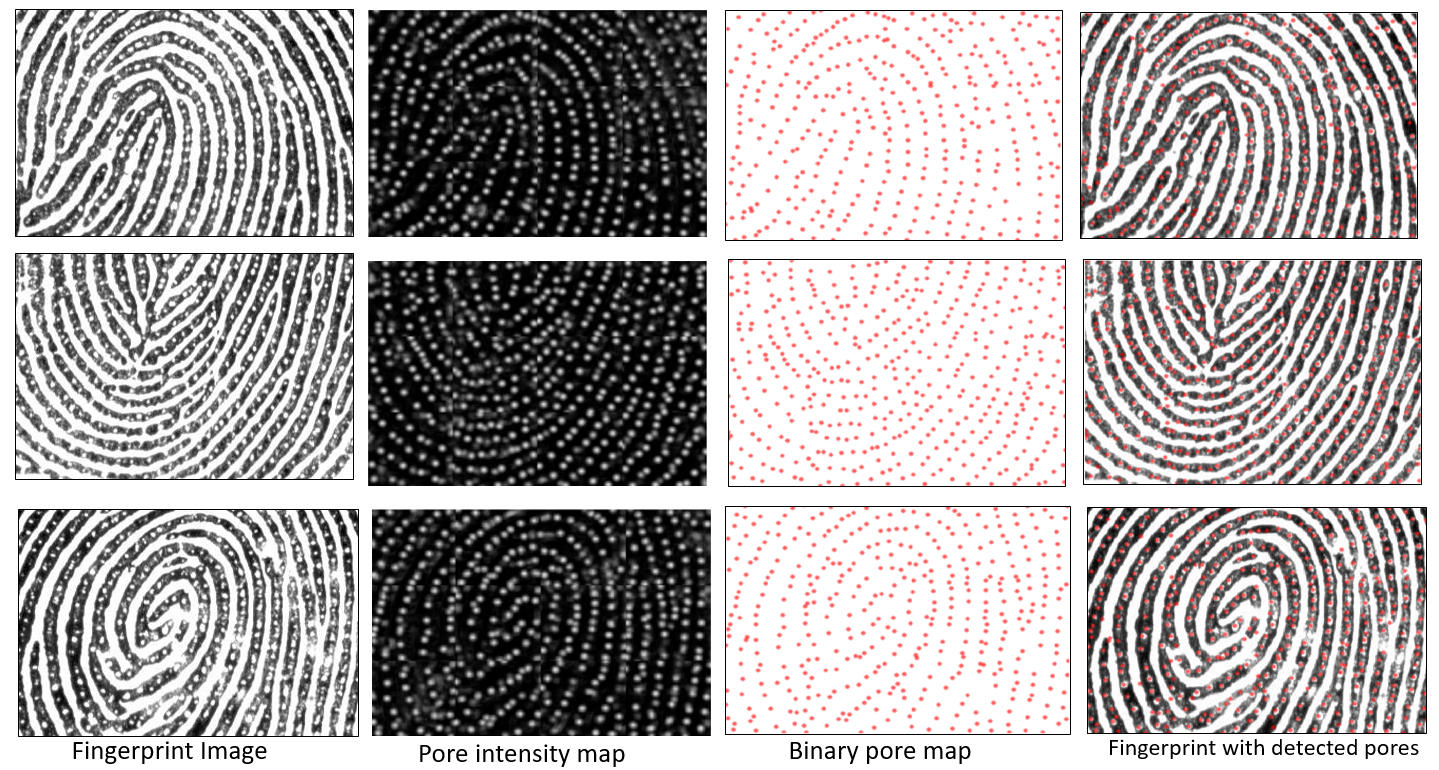}}
\caption{Outputs of different stages of the proposed approach}
\label{pore_detection}
\end{figure*}

  Pore intensity maps generated by DeepDomainPore for test fingerprints are shown in Fig. \ref{pore_detection}. As can be seen, the pores appear clearly as grey-level blobs and rest of the fingerprint features are suppressed. The centers of these grey-level blobs (pores) correspond to the local maxima in the pore intensity map. Thus, we have employed a simple spatial filtering approach to estimate the pore coordinates \cite{vij_cnn}. In this approach, firstly, a maximum value map is obtained by filtering the pore intensity map with a maximum filter of window size $5\times5$ pixels.
Next, for every pixel in the pore intensity map, if its value is equal to the corresponding pixel value in the maximum value map and greater than a pre-defined threshold $th_{p}$, the corresponding pixel in the binary pore map is set to 1. If the aforementioned condition is not satisfied, the corresponding pixel in the binary pore map is set to zero.  Sample fingerprint images from the target domain and their corresponding pore intensity maps and binary pore maps obtained using the proposed approach are presented in Fig. \ref{pore_detection}.

\section{Experimental results and discussion}
In this section, we present the details of how the datasets are prepared to carry out cross-sensor pore detection experiments, followed by the results of our experiments.
\label{results}

\subsection{Dataset preparation}
As discussed previously, all the existing learning-based pore detection approaches have been trained and tested on fingerprint images from PolyU HRF dataset \cite{POLYU}. 
To perform cross-sensor pore detection experiments, we have expanded the in-house IITI-HRF high-resolution fingerprint dataset \cite{Anand2019} to include fingerprint images of 100 subjects. Currently, IITI-HRF high-resolution fingerprint dataset contains fingerprint images of 8 fingers (all fingers except the little fingers of both the hands), each contributing eight impressions. These images are captured using the commercially available Biometrika HiScan-Pro fingerprint scanner.
This scanner captures a fingerprint image of size $1000\times 1000$ pixels. Our dataset, which contains 6400 ($100\times 8 \times 8$) fingerprint images of size $1000\times 1000$ pixels, is referred to as IITI-HRFC. Further, we have created a partial fingerprint image dataset (referred to as IITI-HRFP) by cropping a rectangular region of size $320\times240$ pixels around the center of each image in IITI-HRFC.
The details of IITI-HRF datasets are summarized in Table \ref{IITI_table}.

\begin{table} [!ht]
\caption{Details of the IITI-HRF datasets}
\label{IITI_table}
\centering
\begin{tabular}{|>{\raggedright\arraybackslash}M{1.5cm}|M{1.1cm}|M{1.5cm}|c|c|}
\hline
\multicolumn{1}{|>{\centering\arraybackslash}M{1.5cm}|}{Dataset}&Resolution (dpi)&Image size (pixels)&Fingers&Total images\\
\hline
 IITI-HRFP&1000 &$320\times 240$&800&6400\\
\hline
 IITI-HRFC&1000 &$1000\times 1000$&800&6400\\
\hline
\end{tabular}
\end{table}

Our cross-sensor pore detection experiments have been performed on IITI-HRFP dataset. We have generated ground truth pore data for 20 fingerprints (belonging to 20 unique fingers) selected randomly from IITI-HRFP. These fingerprint images and their corresponding ground truth pore coordinates are collectively referred to as IITI-HRFP-GT. Overall, IITI-HRFP-GT contains 6,406 annotated pores.
Five out of 20 fingerprint images from IITI-HRFP-GT have been used for validation and fine-tuning, while the remaining 15 fingerprint images have been used for testing the proposed DeepDomainPore model. All the impressions of the remaining 780 fingers (having no ground truth data) form the target domain $T$ for training DeepDomainPore model. 
We have divided each fingerprint image in $T$ into non-overlapping patches of size $80\times80$ pixels. Thus, we have a total of 74,880 fingerprint patches from $T$ to train DeepDomainPore model. The authors in \cite{tex_PAMI} have provided pore annotations for 120 fingerprint images of PolyU HRF dataset. Out of 120 fingerprint images of size $640\times480$ pixels, we have selected 90 fingerprint images from the first session to form the source domain $S$ training images. Each image in the source training set is divided into overlapping patches of size $80\times80$ pixels with a step size of 10 pixels. Overall, we have a total of 210,330 fingerprint patches from $S$ to train DeepDomainPore model.

To train DeepDomainPore model, we require  fingerprint images from the source domain $S$ with their corresponding pore label images $I_{P}$ and fingerprint images from the target domain $T$ without pore label images. We have generated the pore label images from the pore annotations. Specifically, the pixels corresponding to the ground truth pore coordinates are marked as 1 in the pore label image, and the pixels present within a radius of 5 pixels from each of the ground truth pore are marked with the value between 0 and 1 and the remaining pixels are marked as 0, in the following manner \cite{Deep_pore,vij_cnn}:
\begin{equation}
\label{pore_label}
  I_{P}(i,j)=\begin{cases}
    1-\frac{d_{pg}(i,j)}{5}, & \text{if $d_{pg}(i,j)<5$}.\\
    0, & \text{otherwise}.
  \end{cases}
\end{equation}
where $I_{P}(i,j)$ is the label assigned to the pixel at $(i,j)$ and $d_{pg}(i,j)$ is the Euclidean distance between the pixel at $(i,j)$ and a ground truth pore.

\subsection{Experimental results}
\label{ER}
We have trained  DeepDomainPore in an end-to-end manner to optimize the pore detection loss $L_{pore}$ and the domain classifier loss $L_{d}$ on the source domain and target domain fingerprint images, respectively. DeepDomainPore has been trained for 10 epochs (each containing 9360 iterations) using a batch size of 8 for both the source and target training sets. The learning rate is set to 0.0001. The number of epochs has been determined based on the training loss and the average true detection rate on the validation set. 
The loss functions have been optimized using adaptive moment estimation (ADAM) \cite{ADAM}. 
The domain adaptation factor $\lambda$  is set to 0.005. 
All our experiments have been performed using PyTorch \cite{paszke2017automatic} in Python environment on a computer with 3.60 GHz Intel core i7-6850K processor, 48 GB RAM, and Nvidia GTX 1080 8 GB GPU.  

The performance of all the approaches has been evaluated using the following metrics: true detection rate ($R_{T}$) and false detection rate ($R_{F}$) \cite{LABATI2017, Deep_pore, vij_cnn}. $R_{T}$ is defined as the ratio of the correctly detected pores to the pores present in the ground truth. On the other hand, $R_{F}$ is defined as the ratio of the falsely detected pores to the total number of detected pores. $R_{T}$ and $R_{F}$ are equivalent to $recall$ and $1-precision$, respectively, of the pore detection approach.  We also report F-score, which is the harmonic mean of $recall$ and $precision$ values.
In addition, we present the receiver operating characteristic  (ROC) curve to ascertain the performance of the proposed approach in cross-sensor scenario.

\begin{table*}[ht!]
\caption{Cross-sensor performance comparison with the existing methods on IITI-HRFP-GT test images}
\centering
\begin{tabular}{|M{3cm}|M{2cm}|M{2cm}|M{2cm}|} \hline
\multirow{2}{*}{Method}&\multicolumn{3}{c|}{Metric}\\ \cline{2-4}
&$R_{T}$& $R_{F}$ & F-score \\ \hline
DeepPore \cite{Deep_pore}&$86.34\%$&$19.11\%$ & $83.23\%$ \\ \hline
DeepResPore \cite{vij_cnn}&$70.91\%$&$19.13\%$ & $74.75\%$ \\ \hline 
Dahia and Segundo \cite{Dahia_pore_detection}&$85.19\%$&$29.09\%$ & $76.79\%$ \\ \hline
DeepResPore-FT &$85.89\%$&$19.10\%$ & $82.70\%$ \\ \hline
DeepDomainPore & \textbf{88.09\%} & $19.17\%$ & \textbf{83.93\%} \\ \hline
\end{tabular}
\label{results_DA}
\end{table*}

\begin{table*}[ht!]
\caption{Cross-sensor performance comparison with the existing methods on the entire set of images from IITI-HRFP-GT}
\centering
\begin{tabular}{|M{3cm}|M{2cm}|M{2cm}|M{2cm}|} \hline
\multirow{2}{*}{Method}&\multicolumn{3}{c|}{Metric}\\ \cline{2-4}
&$R_{T}$& $R_{F}$ & F-score \\ \hline
DeepPore \cite{Deep_pore}&$85.66\%$&$19.47\%$ & $82.77\%$ \\ \hline
DeepResPore \cite{vij_cnn}&$73.98\%$&$19.47\%$ & $76.53\%$ \\ \hline 
Dahia and Segundo \cite{Dahia_pore_detection}&$85.05\%$&$30.59\%$ & $75.92\%$ \\ \hline
DeepDomainPore & \textbf{88.12\%} & $19.47\%$ & \textbf{83.82\%} \\ \hline
\end{tabular}
\label{results_DA_IITI_20}
\end{table*}

To obtain the aforementioned performance metrics, it is important to understand how the detected pores are treated as true pores or falsely detected ones. We have employed the protocol presented in \cite{Dahia_pore_detection}. For a given fingerprint, let the detected pores and the corresponding ground truth pores be represented as  $P_{d}$ and $P_{g}$, respectively. Firstly, we compute the pairwise Euclidean distance between $P_{d}^{i}$ and $P_{g}^{j}$ and store  it in a distance matrix $D_{pg}^{i,j}$, for $i = 1, 2, \hdots, N$ and $j = 1, 2, \hdots, M$ where $N$ and $M$ are the number of detected pores and the ground truth pores, respectively. Next, we find the index of the minimum element of each row of $D_{pg}^{i,j}$. These indices $j^{'}\in \{1, 2, \hdots, M\}$ provide the coordinates of the nearest ground truth pore for each of the detected pores. Further, we find the nearest detected pore for the ground truth pore obtained in the previous step. This is done by examining the $j^{'}th$ column of $D_{pg}^{i,j}$ to find the index $i^{'} \in \{1, 2, \hdots, N\}$ of the minimum element along the column.  Finally, a detected pore $P_{d}^{i}$ is considered to be a true pore only if $ i = i^{'}$, \textit{i.e.} the distance between the detected pore and a ground truth pore is minimum, bidirectionally.

To perform a comparative performance analysis,  we have also evaluated the existing approaches \cite{Deep_pore,vij_cnn, Dahia_pore_detection} on test fingerprint images from IITI-HRFP-GT. The existing approaches \cite{Deep_pore,vij_cnn} have been trained on the same set of 210,330 fingerprint patches from the source domain, while the approach presented in \cite{Dahia_pore_detection} has been trained on a different set of images from the source domain.
We have also evaluated the cross-sensor  pore detection performance of a fine-tuned model referred to as DeepResPore-FT. Specifically, we have fine-tuned DeepResPore model by using the fingerprint images belonging to validation set of IITI-HRFP-GT. In this process, we have retained weight values of the trained DeepResPore model in all the layers except for the last convolutional layer, which has been replaced with a new convolutional layer whose weights are learned using the cross-sensor fingerprint images and pore labels. Each fingerprint image from the validation set of IITI-HRFP-GT is divided into overlapping patches of size $80\times80$ pixels with a step size of 10 pixels. Thus, we have 2,125 patches to fine-tune the model. The pore label images have been generated according to (\ref{pore_label}). We have trained DeepResPore-FT model for 20 epochs with a learning rate of 0.01.

\begin{figure}[!ht]
\centering
{\includegraphics[width=0.5\textwidth, height = 5cm]{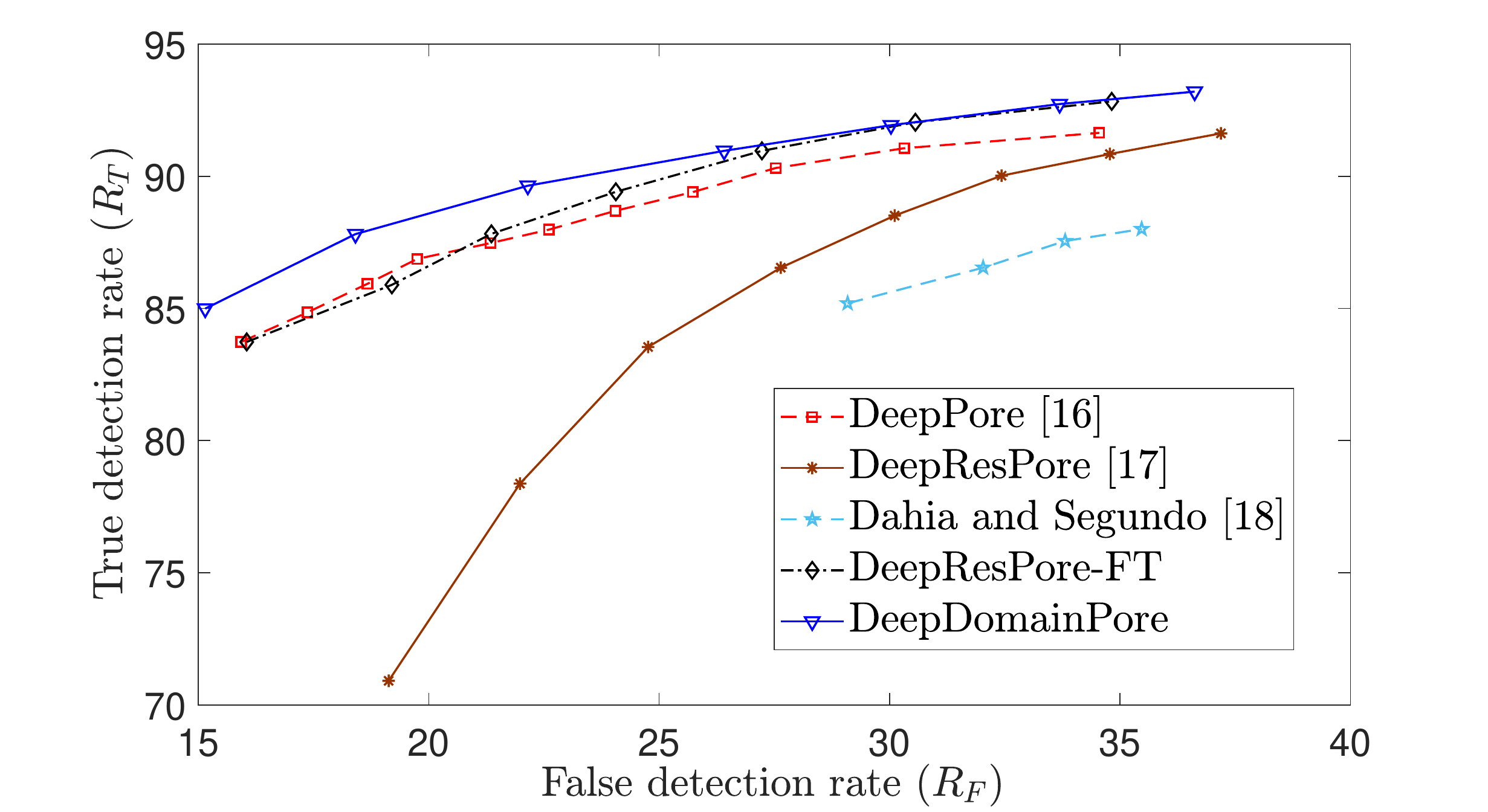}}
\caption{ROC curves in the cross-sensor scenario using 15 test images from IITI-HRFP-GT }
\label{ROC_DA}
\end{figure}

\begin{figure}[!ht]
\centering
{\includegraphics[width=0.5\textwidth, height = 5cm]{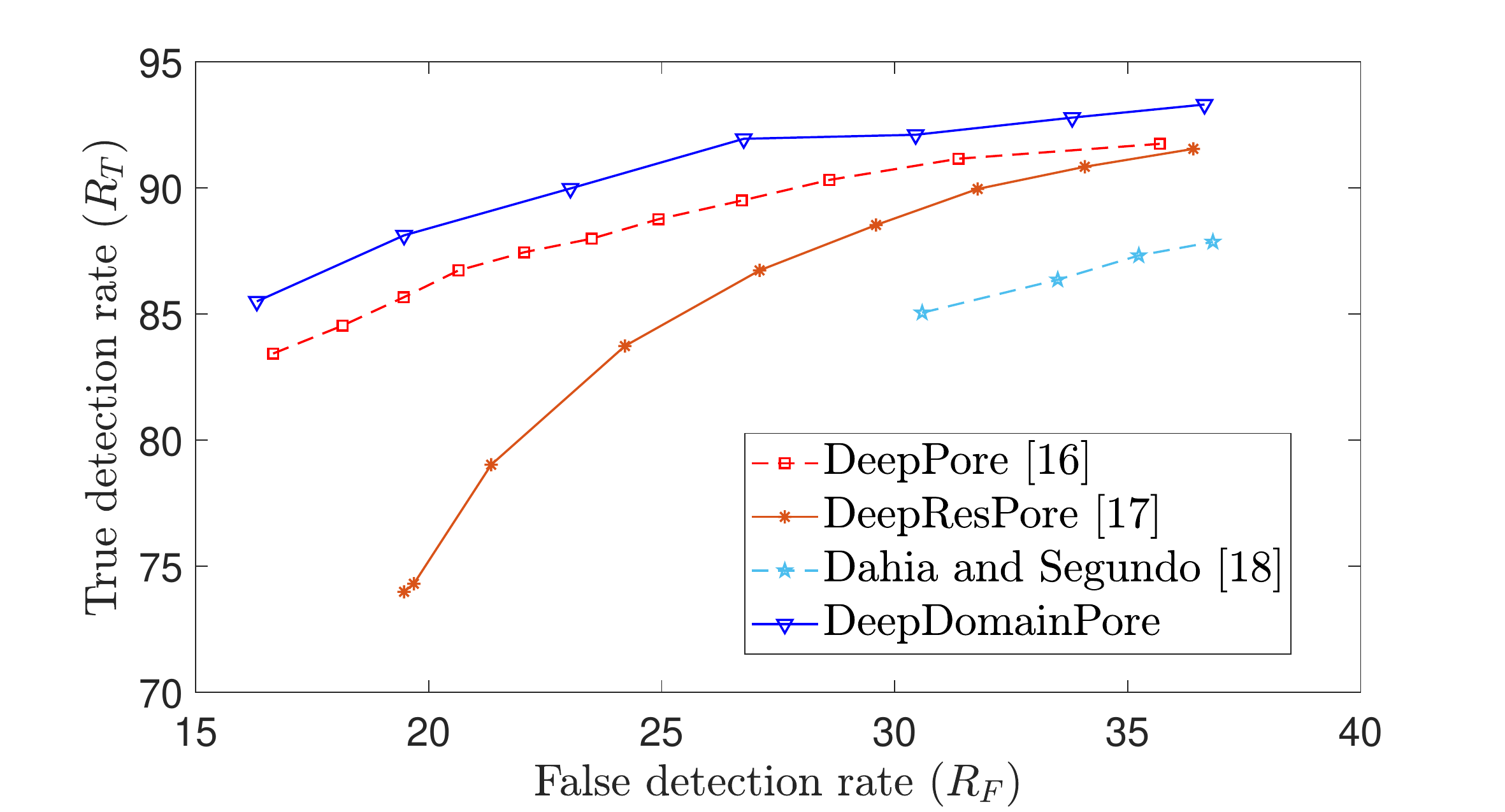}}
\caption{ROC curves in the cross-sensor scenario using the entire set of images from IITI-HRFP-GT }
\label{ROC_DA_IITI_20}
\end{figure}

The results of our experiments in the cross-sensor scenario are presented in Table \ref{results_DA}. To make a fair comparison, $R_{F}$ value has been fixed at approximately 19 and the corresponding $R_{T}$ and F-score values have been computed for the proposed DeepDomainPore, DeepRespore-FT and the existing approaches \cite{Deep_pore, vij_cnn}. In the case of \cite{Dahia_pore_detection}, we have not been able to set $R_{F}$ value to 19, as it detects a large number of false pores resulting in higher value of $R_{F}$. Therefore, we have reported its minimum $R_{F}$ and the corresponding $R_{T}$ and F-score. 
As can be observed in Table \ref{results_DA}, the proposed DeepDomainPore model provides higher $R_{T}$ and F-score values.  
To ascertain this superior performance, we have plotted ROC curves (please see Fig. \ref{ROC_DA}) by varying $th_{p}$ values. 
These curves indicate that the proposed approach provides higher $R_{T}$ at lower values of $R_{F}$, specifically, for $R_{F}$ below 25. Our experimental results clearly indicate that incorporating domain adaption helps improve the pore detection performance considerably in the cross-sensor scenario.  
Further, we have evaluated the methods on the entire set of 20 images from IITI-HRFP-GT. In this set of experiments, we have omitted DeepResPore-FT due to the lack of a separate validation set to fine tune the model. The results of our experiments are presented in Table \ref{results_DA_IITI_20}, and the corresponding ROC curves are presented in Fig. \ref{ROC_DA_IITI_20}.  
These results confirm our earlier observations about the effectiveness of the proposed pore detection method.




\begin{table*}[ht!]
\caption{Performance comparison with the existing methods on the source domain consisting of the entire set of PolyU-HRF-GT images}
\centering
\begin{tabular}{|M{3cm}|M{2cm}|M{2cm}|M{2cm}|} \hline
\multirow{2}{*}{Method}&\multicolumn{3}{c|}{Metric}\\ \cline{2-4}
&$R_{T}$& $R_{F}$ & F-score \\ \hline
DeepPore \cite{Deep_pore}&$84.85\%$&$2.52\%$ & $90.49\%$ \\ \hline
DeepResPore \cite{vij_cnn}&$85.00\%$&$2.50\%$ & $90.44\%$ \\ \hline 
DeepDomainPore & \textbf{87.41\%} & $2.52\%$ & \textbf{92.04\%} \\ \hline
\end{tabular}
\label{results_source}
\end{table*}

\begin{table*}[ht!]
\caption{Performance comparison with the existing methods  on a subset of PolyU-HRF-GT images}
\centering
\begin{tabular}{|M{3cm}|M{2cm}|M{2cm}|M{2cm}|} \hline
\multirow{2}{*}{Method}&\multicolumn{3}{c|}{Metric}\\ \cline{2-4}
&$R_{T}$& $R_{F}$ & F-score \\ \hline
DeepPore \cite{Deep_pore}&$93.19\%$&$8.89\%$ & $92.10\%$ \\ \hline
DeepResPore \cite{vij_cnn}&$93.01\%$&$8.89\%$ & $91.98\%$ \\ \hline 
Dahia and Segundo \cite{Dahia_pore_detection}&$91.95\%$&$8.88\%$ & $91.53\%$ \\ \hline
DeepDomainPore & \textbf{94.55\%} & $8.88\%$ & \textbf{92.77\%} \\ \hline
\end{tabular}
\label{results_source_last10}
\end{table*}

\begin{figure}[!ht]
\centering
{\includegraphics[width=0.5\textwidth, height = 5cm]{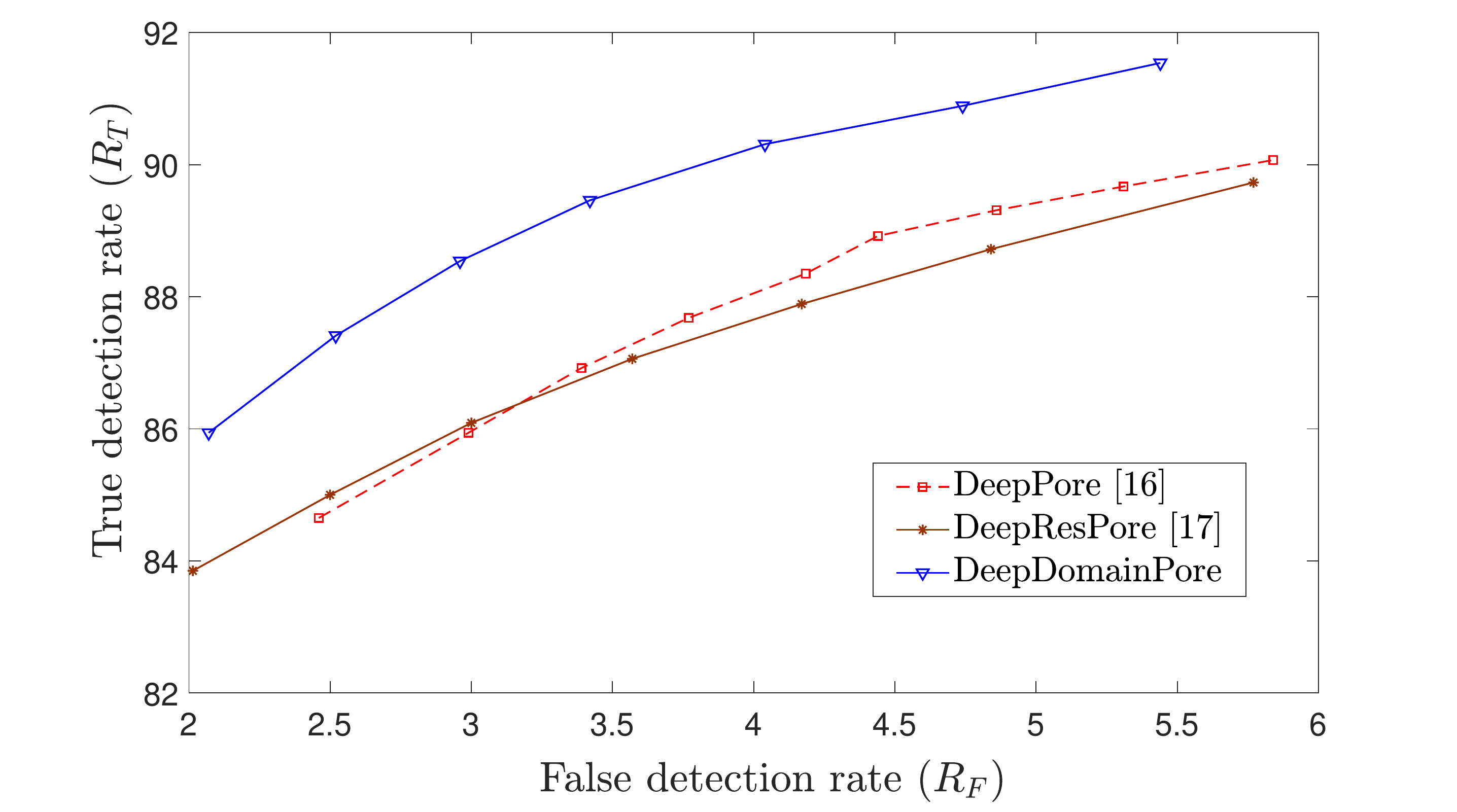}}
\caption{A comparative analysis using ROC curves on source domain}
\label{ROC_Polyu}
\end{figure}

Further, we have compared the performance of DeepDomainPore with the existing approaches \cite{Deep_pore, vij_cnn} on fingerprint images belonging to the source domain. We refer to the entire set of 30 fingerprint images, for which ground truth pore coordinates have been provided in PolyU HRF dataset \cite{POLYU}, as PolyU-HRF-GT. 
In this set of experiments, the fingerprint images from PolyU-HRF-GT, which is independent of the set of source domain images used for training the models, form the test set. Table \ref{results_source} presents the results obtained when $R_{F}$ is fixed at approximately 2.5 and the corresponding $R_{T}$ and F-score values are computed for the existing and the proposed approaches. As can be seen, the proposed approach achieves higher $R_{T}$ and F-score compared to the current state-of-the-art approaches on a test set from the source domain as well.
ROC curves shown in Fig. \ref{ROC_Polyu} clearly indicate that the proposed DeepDomainPore model achieves consistently higher $R_{T}$. Since the approach presented by Dahia and segundo \cite{Dahia_pore_detection} uses a part of PolyU-HRF-GT images for training, their model could not be tested on the entire set of PolyU-HRF-GT images. Therefore, its performance has not been not reported in Table \ref{results_source}. However, we have selected the last 10 images from PolyU-HRF-GT (as was done by the authors in \cite{Dahia_pore_detection}) to form a test set and carried out another set of  experiments, results of which are reported in Table \ref{results_source_last10}. For a fair comparison of results, $R_{F}$ value has been set to 8.88 (as was reported previously in \cite{Dahia_pore_detection}) and the corresponding $R_{T}$ and F-score values have been computed for the proposed and the existing approaches \cite{Deep_pore,vij_cnn}. As can be seen, the proposed DeepDomainPore based pore detection approach consistently provides higher $R_{T}$ and F-score.

Overall, the experimental results presented in this section indicate that the proposed DeepDomainPore model provides improvement in performance over the current state-of-the-art approaches for both the cross-sensor and the source domain scenarios. Specifically, it achieves higher $R_{T}$ for low $R_{F}$ in both the scenarios considered in this work.

\section{Conclusion}
\label{conclude}
In this paper, we have presented a learning-based methodology for detection of pores in high-resolution fingerprint images. The proposed CNN model, DeepDomainPore, has been specifically designed to handle cross-sensor scenarios. DeepDomainPore is a combination of a residual learning-based CNN and the unsupervised domain adaptation, incorporated by embedding a gradient reversal layer between the CNN and the domain classifier network. 
Our experimental results on cross-sensor fingerprint images demonstrate the effectiveness of the proposed DeepDomainPore in detecting pores in high-resolution fingerprint images with a significant domain shift.
Most importantly, the proposed DeepDomainPore model provides state-of-the-art performance on test sets from both the source (PolyU-HRF-GT) and the target (IITI-HRFP-GT) domains. Specifically, the proposed approach correctly detects 5,644 out of 6,404 pores present in the high-resolution fingerprint images belonging to the target domain.  


\ifCLASSOPTIONcaptionsoff
  \newpage
\fi



%



\bibliographystyle{IEEEtran}
\bibliography{IEEEabrv,vijay_pore_match}

%








\end{document}